# Degradation-Agnostic Statistical Facial Feature Transformation for Blind Face Restoration in Adverse Weather Conditions

Chang-Hwan Son* and Cheol-Hwan Kim

Department of Software Science & Engineering, Kunsan National University

558 Daehak-ro, Gunsan-si 54150, Republic of Korea

*Corresponding Author

Phone Number: 82-63-469-8915; Fax Number: 82-63-469-7432

E-MAIL: cson@kunsan.ac.kr

**Abstract**

Adverse weather severely degrades face images through complex effects such as rain streaks, haze effects, blurring, and resolution loss, making recognition highly unreliable. Existing blind face image restoration (FIR) methods remain limited because they lack modules dedicated to handling weather-induced degradations. We propose a blind FIR framework based on degradation-agnostic statistical facial feature transformation, integrating two components: Local Statistical Facial Feature Transformation (SFFT) and Degradation-Agnostic Feature Embedding (DAFE). The local SFFT module enhances facial structure and color fidelity by aligning the local statistical distributions of low-quality (LQ) facial regions with those of high-quality (HQ) counterparts, ensuring that critical facial components such as the eyes, nose, and mouth are sufficiently restored for face recognition under adverse weather. The DAFE module further enables robust statistical facial feature extraction by aligning LQ and HQ encoder representations, making the restoration process adaptive to severe weather-induced degradations. Experiments show that the proposed degradation-agnostic SFFT model restores key facial components and outperforms state-of-the-art FIR methods, demonstrating its potential to enable reliable face recognition under adverse weather conditions.

## 1. Introduction

Face image restoration (FIR) aims to recover high-quality (HQ) images from degraded inputs. With the advancement of intelligent CCTV systems, FIR techniques are being actively developed to enhance face recognition and analysis, particularly in applications such as missing person identification, suspect detection, and visitor management. Conventional FIR methods [1] primarily focus on super-resolving low-resolution faces under the assumption that degradation parameters, such as noise level and downsampling ratio, are known a priori. However, this assumption does not hold in real-world scenarios. To this end, blind FIR approaches [2] that do not require explicit knowledge of degradation parameters have been introduced.

Generative Adversarial Networks (GANs) [3] have emerged as a core framework for blind FIR, as they effectively model the probability density distribution of face samples through adversarial learning. GAN-based blind FIR approaches leverage various priors, including geometric [4,5], reference [6,7,8], and generative priors, [2,9,10] to improve facial details and generate faithful faces. These advancements enable the restoration of low-quality (LQ) face images affected by complex degradations, thereby improving the robustness and applicability of FIR techniques in real-world conditions.

Nevertheless, recent blind-FIR methods have certain limitations. Intelligent CCTVs are typically deployed in outdoor environments, where captured images are highly susceptible to severe degradations under adverse weather conditions, such as heavy rain and deep fog [11,12]. While conventional FIR methods can be employed as an alternative to restore the LQ facial images affected by such conditions, their performance remains suboptimal, as existing FIR methods do not incorporate dedicated restoration modules that explicitly address weather-induced degradations. Consequently, facial textures and structures may become distorted, and weather-related artifacts, such as rain streaks and fog, may not be effectively removed.

***Fig. 1 shows examples where an LQ facial image captured under heavy rain is extremely degraded due to multiple degradation processes, including rain streaks, blurring, and downsampling, making the face visually unrecognizable and thereby hindering face recognition. Although conventional FIR models such as GANs and diffusion models have been applied, they fail to restore critical facial components and to show the potential to enable reliable face recognition.***

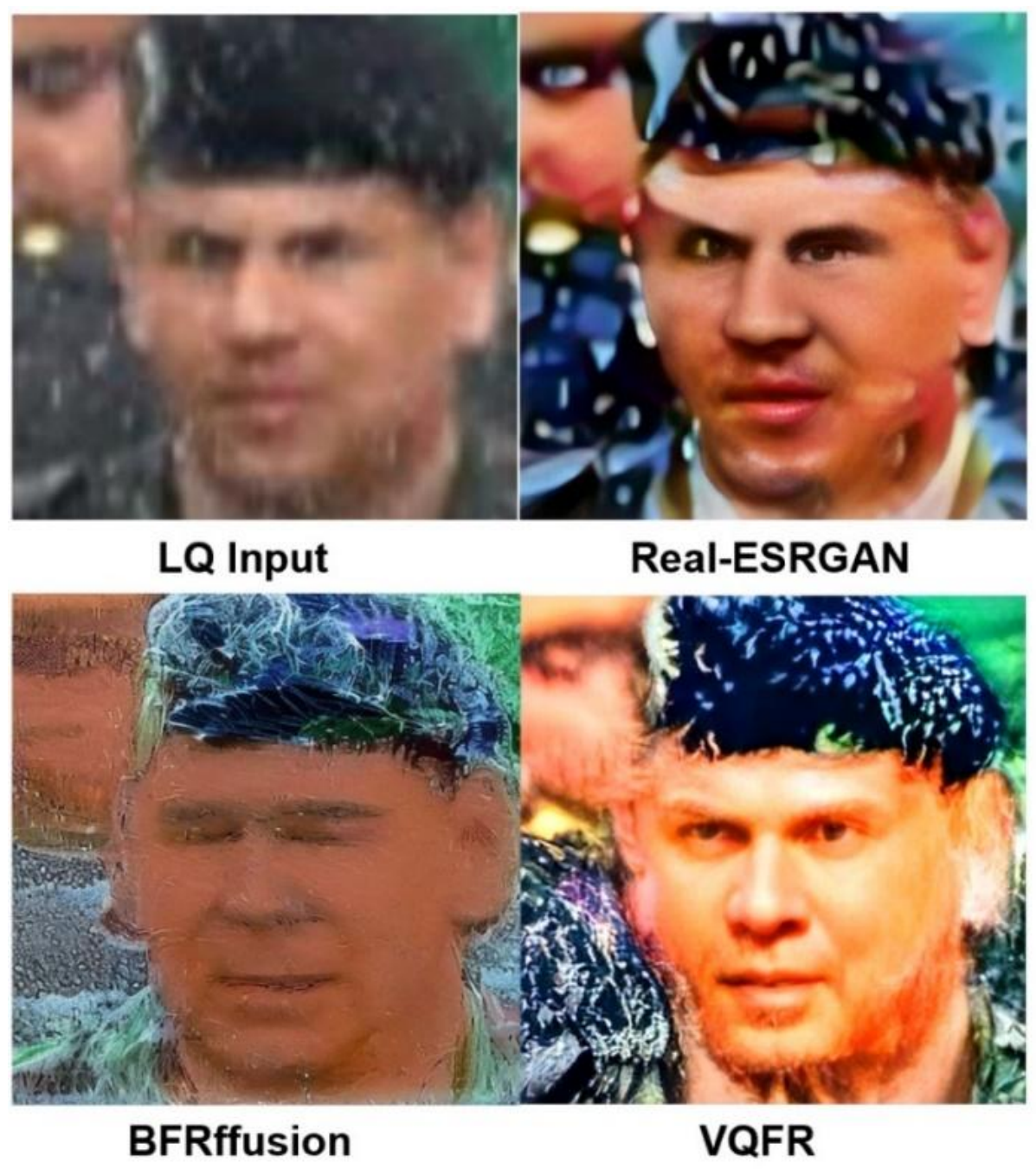

Fig. 1. Limitations of existing FIR models on facial images severely degraded by adverse weather.

***To the best of our knowledge, research on blind FIR methods specifically designed to overcome the severe degradations caused by adverse weather conditions and to address the challenges of reliable face recognition remains scarce. As shown in Fig. 1, recently introduced state-of-the-art FIR models based on diffusion and GAN frameworks still fail to restore facial components sufficiently for face recognition under adverse weather conditions.*** Therefore, in this study, we propose a novel blind FIR method to recover critical facial components and enable face recognition from visually indiscernible LQ images affected by adverse

weather conditions. Two primary contributions are introduced. The first is the Statistical Facial Feature Transformation (SFFT), a significant advancement over the statistical feature transformation (SFT) [13] that captures and manipulates statistical distributions of facial components. SFT transforms target statistics, characterized by mean and variance, into reference statistics. Originally, it was applied in the image domain for color transfer [14], aiming to adjust the color appearance of a target image to match that of a reference image. Recently, SFT has been adopted in the feature domain for blind FIR [5] within the GAN framework, where it maps the statistics of LQ facial features into those of HQ facial features. Although this model is one of the state-of-the-art FIR methods, it lacks a dedicated weather-aware restoration process, and its performance is therefore limited. Moreover, in this model, SFT was applied to the entire face, focusing solely on global statistical transformation. Consequently, the detailed representation of facial components is insufficient. Since face images are highly structured data, a more optimal approach, namely SFFT, is required to incorporate local SFT and achieve more accurate and precise facial representation.

The second is the Degradation-Agnostic Feature Encoding (DAFE), designed to enhance the robustness of feature extraction, regardless of weather-induced degradations. Recently, vision-language pretraining, e.g., CLIP [15] and ALIGN [16], has emerged as a promising alternative for visual representation learning. The core idea is to align multimodal image and text features using two separate encoders via contrastive loss, which pulls together positive text-image pairs while pushing apart negative ones in the feature space. Inspired by this approach, this study employs two image encoders: One for HQ face images and another for LQ face images. Unlike conventional vision-language pretraining, which utilizes text and image encoders, this study aligns image features extracted from HQ and LQ encoders. During the training phase, the HQ encoder is used, whereas during inference, the pretrained LQ encoder is leveraged to extract HQ-like features from LQ face input. The ultimate objective of utilizing these encoders is to derive the statistical distribution from the HQ features, thereby enabling the proposed SFFT to achieve more precise and refined transformation. It is important to emphasize that the role of the image encoders is to enhance the SFFT rather than merely performing feature matching. A similar approach [12] using image encoders has been introduced for image captioning under heavy rain conditions. However, a key distinction exists: in the image captioning model, two encoders were employed solely for feature matching, whereas in this study, the primary focus is on SFFT modeling. Furthermore, the ultimate objective of this study is FIR rather than image captioning, and the loss functions used differ.

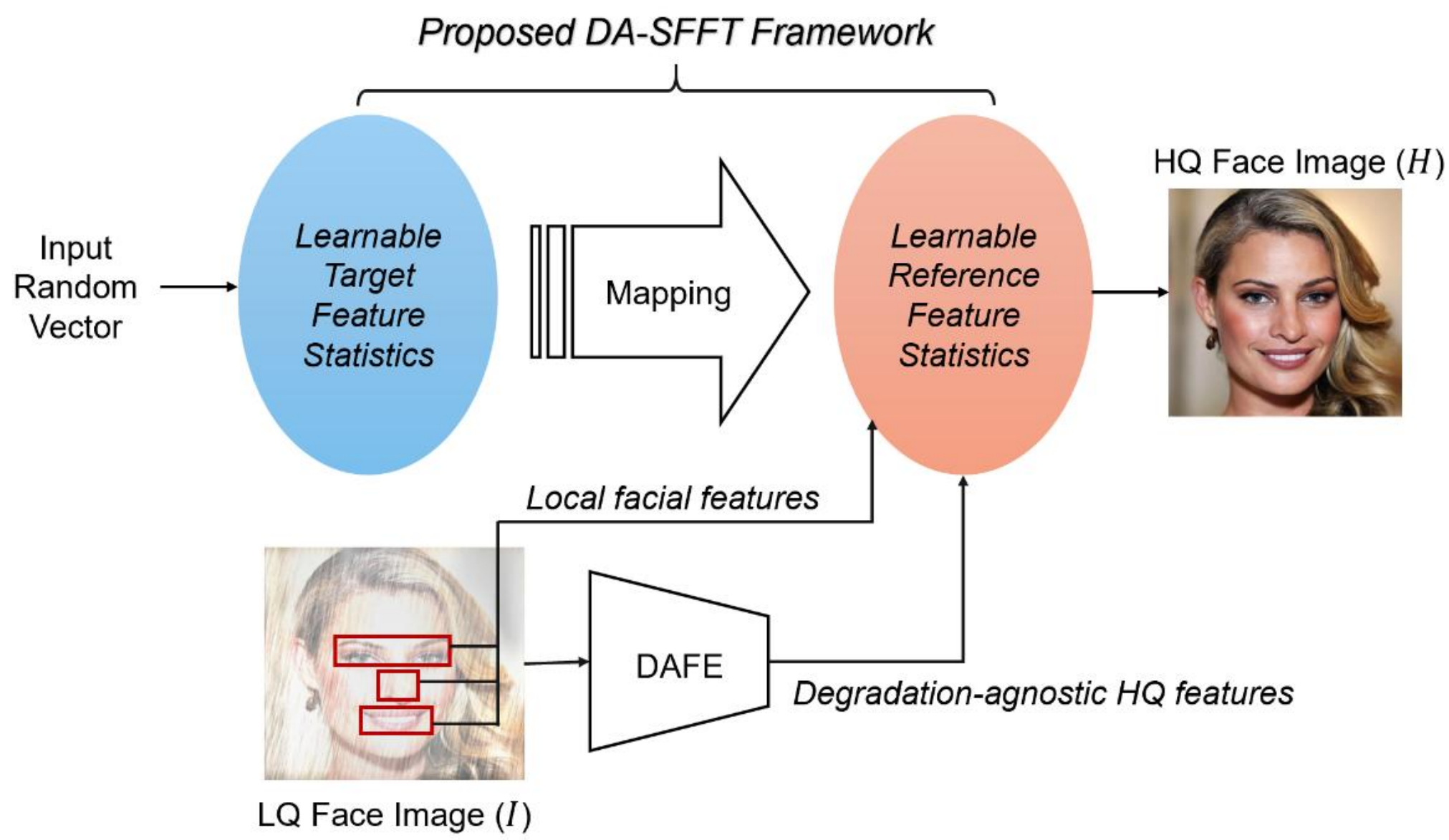

Fig. 2. Concept of the proposed DA-SFFT framework for blind FIR under adverse weather conditions.

Fig. 2 illustrates the concept of the proposed DA-SFFT-based GAN framework. Our goal is to generate HQ face images from input random vectors through adversarial learning. To achieve this, the DA-SFFT is employed to map the target feature statistics to the reference statistics, where the statistics refer to the learnable mean and variance. The key challenge lies in accurately learning the reference statistics, since LQ face images are severely degraded and it is inefficient and unreasonable to directly derive reference statistics from the LQ conditional input. To address this issue, the DAFE module is introduced to extract HQ-like features and transform them into degradation-agnostic reference statistics under various weather conditions. Furthermore, to better reflect local statistics for facial components such as the eyes, nose, and lips, the local SFFT is employed, which is critical for accurately restoring facial components and enabling reliable face recognition under adverse weather conditions.

Our contributions are twofold.

- First, this study introduces a new GAN-based blind FIR specifically designed to address weather-induced degradations. Unlike conventional blind FIR approaches, which primarily focus on restoring LQ images captured under clear weather conditions, this study explicitly accounts for the weather-induced artifacts to enhance face recognition performance. ***With the increasing deployment of intelligent CCTVs, developing a weather-aware restoration process has become crucial. To the best of our knowledge, this is the first attempt to explore FIR under adverse weather conditions and to demonstrate the potential of the proposed model for real-world FIR in such challenging environments.***
- Second, this study proposes a novel DA-SFFT (Degradation-Agnostic Statistical Facial Feature Transformation) framework that integrates the SFFT and DAFE modules for FIR under adverse weather conditions. The proposed SFFT learns the local statistical distributions from LQ face images and transforms them to align with those of HQ face images, thereby improving facial representation. Meanwhile, DAFE leverages a pretrained image encoder to extract *degradation-agnostic* HQ features and derive their statistical distributions. This integration enables SFFT to achieve more precise and effective facial feature transformations. ***Experimental results confirm that the proposed blind DA-SFFT FIR model is capable of restoring critical facial components from visually unrecognizable facial images, thereby enabling robust face recognition under adverse weather conditions.*** Furthermore, the proposed model outperforms state-of-the-art FIR approaches, demonstrating the effectiveness of both the DAFE and SFFT modules for blind facial restoration.

The rest of the paper is organized as follows. Section 2 presents the literature review, and Section 3 introduces our proposed methodology in more detail. Section 4 provides the experimental results with quantitative and qualitative evaluations. Finally, Section 5 concludes our research findings.

## 2. Related Work

### 2.1. Face image restoration

Traditional face restoration methods adopt a non-blind approach [17-21], which assumes that degradation parameters such as down-sampling ratio and noise level are known. However, in real-world applications, this assumption is no longer valid. Consequently, blind face restoration, which aims to restore HQ faces from LQ counterparts suffering from unknown degradation, has been actively studied in recent years. Previous studies have reported that priors play a crucial role in blind face image restoration and are typically categorized into three types: geometric priors [4,5,18,21,22], generative priors [2,9,10,23], and reference priors [6,7,8,24].

Geometric priors leverage face-specific information such as facial landmarks [4,18,22], parsing maps [4,5], and facial component heatmaps [21] to reconstruct accurate facial shapes and details. However, since these priors are estimated from LQ images, their performance is inevitably restricted. To mitigate the dependency on degraded inputs, reference prior-based approaches have been proposed, requiring reference images of the same identity [8] or facial dictionaries [6,7]. However, the reference images are not always accessible, and the limited size of the reconstruction-oriented [6] and recognition-oriented [24] dictionaries restrict their diversity and richness of facial details. Generative priors encapsulated in pretrained models such as StyleGAN [13] have also been exploited for blind face restoration. Early models such as PULSE [10] and mGANPrior [9] find the latent code of high-quality faces via iterative optimization. However, these methods are computationally expensive at inference and often yield suboptimal results because the low-dimensional latent codes are insufficient to guide the restoration. GPEN [25] and GFP-GAN [2] more efficiently incorporate generative priors into the restoration process.

For the past few years, GANs have been the dominant approach in generative modeling. However, recent advances in diffusion models [26-30] have led to groundbreaking developments in computer vision, making them a rapidly evolving alternative to GANs. DR2 [26] proposes a solution for generating degradation-invariant intermediate images to avoid dependency on image degradation models. However, this method requires an additional enhancement module to reconstruct the final HQ images, and its performance depends on control parameters. DiffBIR [27] treats BIR as a conditional image generation problem and proposes a restoration module that creates more reliable conditional images from LQ ones. In the generation module, a variant of ControlNet [28] was used to leverage the prior knowledge of pretrained Stable Diffusion. DiffBFR [29] introduces a cascaded diffusion model with truncated sampling, which starts from LQ images with partial noise added and theoretically shrinks the evidence lower bound of the diffusion probabilistic model. This alleviates the long tail distribution problem. Despite achieving realistic restoration results, diffusion-based methods [30] face the common issue of slow sampling speeds.

Existing blind FIR methods based on GAN and diffusion frameworks, as mentioned above, have mainly addressed LQ face images captured under normal weather conditions, where degradations are limited to downsampling, blurring, and noise. With the increasing use of intelligent surveillance cameras, however, face images are often severely degraded by adverse weather conditions, introducing challenges such as rain streaks, haze, and heavy rainfall. To date, blind FIR methods specifically designed to handle such severe weather-induced degradations remain scarce. In this study, we propose a novel blind FIR method capable of handling more complex degradation models associated with adverse weather conditions, while enabling face recognition from visually indiscernible LQ images under such conditions, representing a clear departure from conventional FIR approaches.

### 2.2. Image restoration under adverse weather conditions

In adverse weather conditions, image restoration (IR) models can be broadly categorized into single-task and all-in-one models. Single-task IR models are tailored to handle a specific type of image degradation—such as rain streaks [31], heavy rain [32], snow [33], or haze [34]—by addressing each degradation type independently. Early IR models were predominantly based on Convolutional Neural Networks (CNNs), which utilize small filters to extract deep features and enhance feature representation. Variants of CNNs incorporating residual learning [31] and multiscale learning [35] have been introduced to improve high-frequency detail preservation and accommodate different scales. Furthermore, attention mechanisms have been integrated into CNNs to emphasize regions of interest (ROIs) and enhance feature discrimination [36]. More recently, Vision Transformers (ViTs) [37], another class of neural architecture, have emerged to capture long-range pixel dependencies based on multi-head attention and feed-forward layers. Additionally, Generative Adversarial Networks (GANs) have been employed as loss supervision mechanisms to enforce solutions that better align with natural image manifolds [32,38-40].

Meanwhile, all-in-one IR models, also referred to as unified models, are capable of restoring LQ images affected by various unknown degradation types and varying levels of severity. These models have attracted growing research attention in recent years. Approaches such as visual prompt learning [41], leveraging vision-language pretraining models like CLIP [15], degradation classifier learning [42], and contrastive learning [43] have been incorporated into IR networks. Notable architectures include U-Net-based vision transformers and diffusion models, which have demonstrated promising advancements in unified image restoration.

Although all-in-one IR models can handle certain weather-induced degradations such as rain or haze, there are key differences compared to the proposed FIR method. All-in-one IR models typically assume that each input image contains a single type of artifact—for example, rain images are assumed to have only rain streaks, without considering rain accumulation (i.e., haze-like visual effects). Consequently, more complex combinations of degradations, such as blurring, downsampling, and noise, are generally excluded. While these models can extract discriminative features from degraded images that are easily discernible by human eyes, this assumption does not hold in real-world scenarios, where LQ images often contain blended degradations that cannot be separated. In contrast, the proposed FIR model is designed to handle multiple degradations, including noise, blurring, downsampling, rain, and haze effects, making it fundamentally different from conventional all-in-one IR approaches. Moreover, to the best of our knowledge, all-in-one IR models have not been applied to FIR under adverse weather conditions.

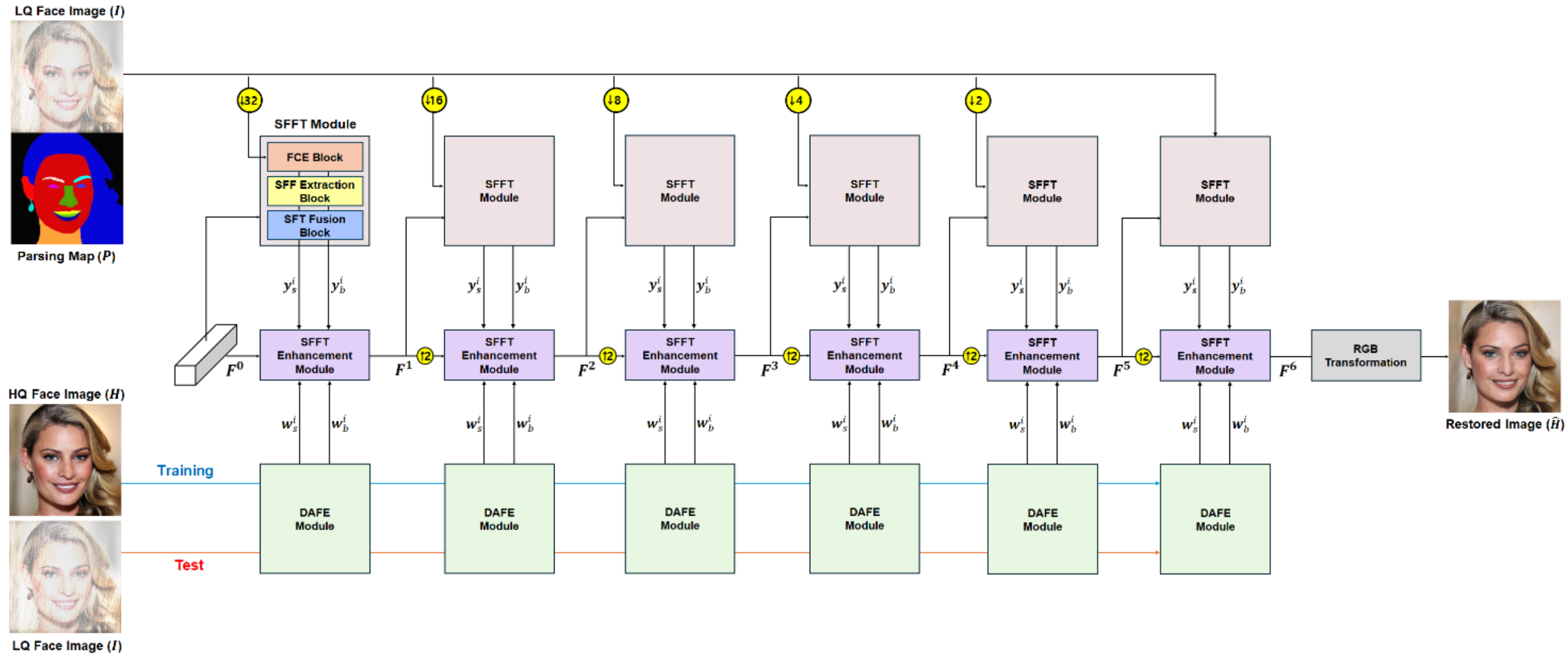

Fig. 3. The proposed DA-SFFT framework for blind FIR under adverse weather conditions.

## 3. Methodology

### 3.1. Overview of DA-SFFT

We describe the proposed DA-SFFT framework for FIR in this section. Given an input LQ face image ($\boldsymbol{I}$) suffering from unknown degradations under adverse weather conditions, the objective of FIR is to reconstruct a HQ image $\widehat{\boldsymbol{H}}$, which closely approximates the ground-truth $\boldsymbol{H}$, in terms of realism and fidelity.

Fig. 3 depicts the proposed DA-SFFT model for blind FIR under adverse weather conditions. The DA-SFFT framework consists of three primary modules: SFFT, DAFE, and SFFT Enhancement modules. First, the SFFT module learns local *statistical facial features (SFF)*—specifically, the mean and standard deviation—corresponding to distinct facial components such as the eyes, nose, and lips. **This allows the local statistical distributions of LQ facial components to match those of their HQ counterparts, thereby enhancing local facial feature representation.**

Second, the **DAFE module aims to generate degradation-agnostic SFF of the ground-truth $\boldsymbol{H}$, regardless of adverse weather conditions**. However, given that adverse weather conditions often degrade input images, directly extracting those features from LQ facial images is inherently challenging. To address this, the proposed approach employs two image encoders that embed features into a shared latent space: one for HQ images and the other for LQ images. These encoders are aligned through a loss function minimization, enabling the pretrained LQ image encoder to effectively extract degradation-agnostic HQ-SFF from LQ inputs during inference. Note that HQ face images are used during training, while LQ face images are used during inference.

Third, the SFFT Enhancement module fuses the SFFs obtained from both the SFFT and DAFE modules to refine and complement the statistical feature representation. The three aforementioned modules operate at different image scales, as depicted in Fig. 3, ensuring a scale-adaptive SFT restoration process.

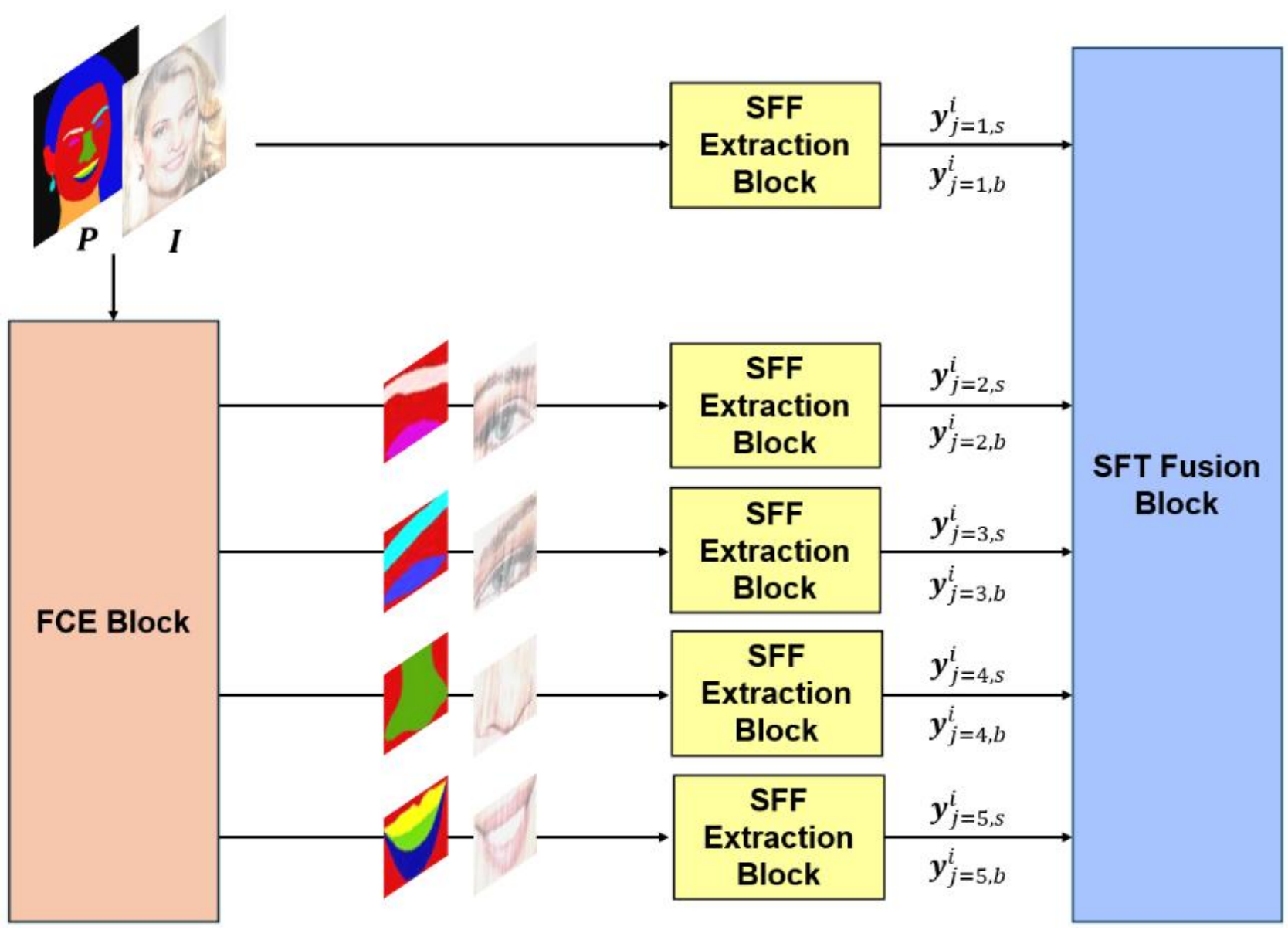

Fig. 4. Structure overview of the SFFT module.

### 3.2. SFFT Module

To extract SFFs from an LQ face image (***I***) and its corresponding parsing map (***P***), and to fuse them for SFT, the SFFT module sequentially performs a series of blocks: facial component extraction (FCE), SFF Extraction, and SFT Fusion blocks, as illustrated in Fig. 4.

#### 3.2.1 FCE Block

The FCE block extracts local facial regions, such as the eyes, nose, and lips, from inputs ***I*** and ***P***. In this study, a straightforward approach based on predefined bounding box coordinates was adopted for cropping local regions. Fig. 4 presents examples of the extracted facial regions. As shown in this figure, different local facial components exhibit distinct their own statistical distributions. Consequently, employing a local SFT is a more optimal approach for enhancing facial representation compared to global SFT, which relies on the statistical distribution of the entire face.

### 3.2.2. SFF Extraction Block

The SFF Extraction block learns statistical parameters, referred to as SFFs, for each facial region as well as for the entire LQ face image.

$$[\boldsymbol{y}_{j,s}^{i}, \boldsymbol{y}_{j,b}^{i}] = \psi_{SFF}(\boldsymbol{I}_{j}^{i}, \boldsymbol{P}_{j}^{i}) \tag{1}$$

Here, $\psi_{SFF}$ denotes the SFF Extraction function. The index $i$ represents the i$th$ SFF Extraction block in Fig. 3, which corresponds to a specific image scale, while the index $j$ denotes the j$th$ facial component, including the entire facial region, left eye, right eye, nose, and mouth. $I_j^i$ and $P_j^i$ represent the j$th$ facial region and its corresponding parsing map at the i$th$ image scale, respectively. $\psi_{SFF}$ outputs two learnable SFFs, $\boldsymbol{y}_{j,s}^{i}, \boldsymbol{y}_{j,b}^{i}$, which are subsequently used as local reference statistics for the application of local SFT. To implement $\psi_{SFF}$, Convolution (Conv) and Rectified Linear Unit (ReLU) layers were utilized. Notably, these parameters are vectors of size $C$, determined by the number of filters in the Conv layer.

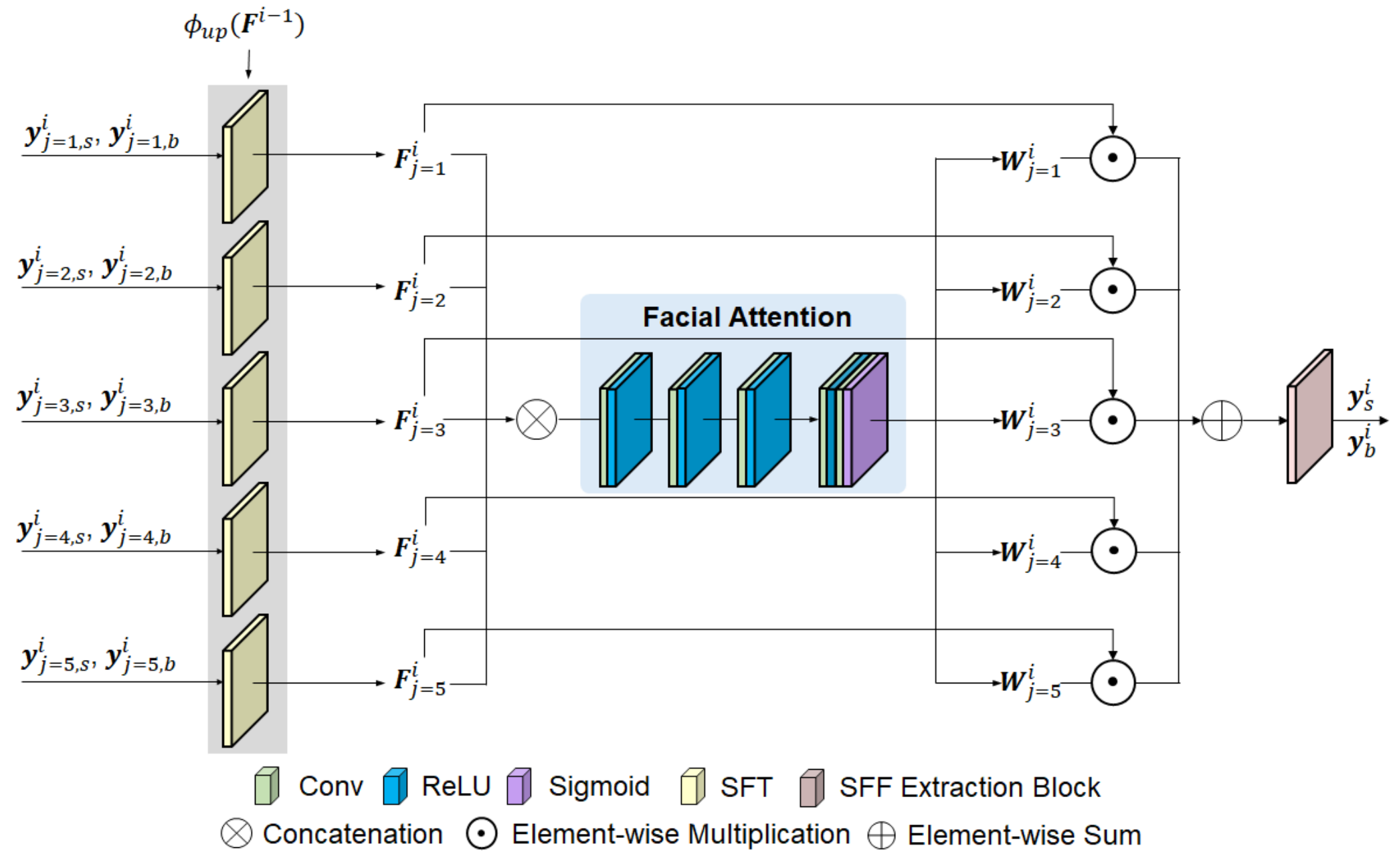

Fig. 5. The architecture of the SFT Fusion block.

### 3.2.3. SFT Fusion Block

Since each facial region has its own SFFs, denoted as $\boldsymbol{y}^i_{j,s}$ and $\boldsymbol{y}^i_{j,b}$, it is essential to determine the relative importance of each SFF and derive the final SFF from the given set of local SFFs. To achieve this, the SFT Fusion block is designed. In this block, the SFT layer is first applied to the input feature map $F^{i-1}$ using the local SFFs, and then the resulting facial feature maps $F^i_j$ are fused using facial attention to generate the unified SFF, denoted as $\boldsymbol{y}^i_s$ and $\boldsymbol{y}^i_b$. The detailed architecture of the SFT Fusion block is illustrated in Fig. 5.

Given the local SFFs $\boldsymbol{y}^i_{j,s}$ and $\boldsymbol{y}^i_{j,b}$, each SFT layer takes the same feature map $\phi(F^{i-1})$ as input, where $F^{i-1}$ initially starts as a random vector and expands in size after passing through the upsampling layer $\phi$ and this feature map corresponds to the output of the SFFT Enhancement module, as depicted in Fig. 3. The SFT layer operates as follows.

$$F^i_j = \boldsymbol{y}^i_{j,s} \frac{\phi(F^{i-1}) - \mu\left(\phi(F^{i-1})\right)}{\sigma\left(\phi(F^{i-1})\right)} + \boldsymbol{y}^i_{j,b} \tag{2}$$

where $\mu$ and $\sigma$ denote the functions that compute the mean and standard deviation of input feature map $\phi(F^{i-1})$. In Eq. (2), $\boldsymbol{y}_{j,s}^{i}$ and $\boldsymbol{y}_{j,b}^{i}$ serve as local reference statistics for each facial component, while $\mu$ and $\sigma$ correspond to the target statistics. The SFT layer begins by normalizing the input feature map $\phi(F^{i-1})$ by subtracting its mean $\mu$ and dividing it by its standard deviation $\sigma$. Subsequently, the normalized feature map is transformed to match the local reference statistics, ensuring statistical consistency with the desired local distribution.

Given the facial feature maps $F_j^i$, it is necessary to determine their relative significance. To this end, facial attention is employed to learn weighting maps $W_j^i$, which capture the relative importance of each facial feature map. As illustrated in Fig. 5, facial attention takes as input the concatenated facial feature maps $F_j^i$ and outputs the learned weighting maps $W_j^i$.

$$W_j^i = FA\left(Concat(F_{j=1}^i, F_{j=2}^i, .., F_{j=N}^i)\right) \tag{3}$$

Here, $FA$ represents the facial attention mechanism responsible for learning the weighting maps $W_j^i$, while $N$ denotes the number of facial components, and $Concat$ represents the concatenation layer. The facial attention is constructed using convolution blocks (Conv+ReLU), and a sigmoid layer is added in the final block to squash the values into the range [0, 1].

To determine the unified reference statistics, the SFF Extraction is applied as follows:

$$[\boldsymbol{y}_s^i,\ \boldsymbol{y}_b^i] = \psi_{SFF}\left(\textstyle\sum_{j=1}^{N} F_j^i \odot W_j^i\right) \tag{4}$$

where $\odot$ indicates element-wise multiplication. In Eq. (4), each facial feature map $F_j^i$ is weighted by its corresponding attention map $W_j^i$, and the weighted maps are aggregated into a unified facial feature map. The SFF Extraction block $\psi_{SFF}$ then generates the final SFF, which serves as reference statistics. Consequently, as illustrated in Fig. 5***, the SFT Fusion block integrates local SFFs as input and produces the unified SFF as output.***

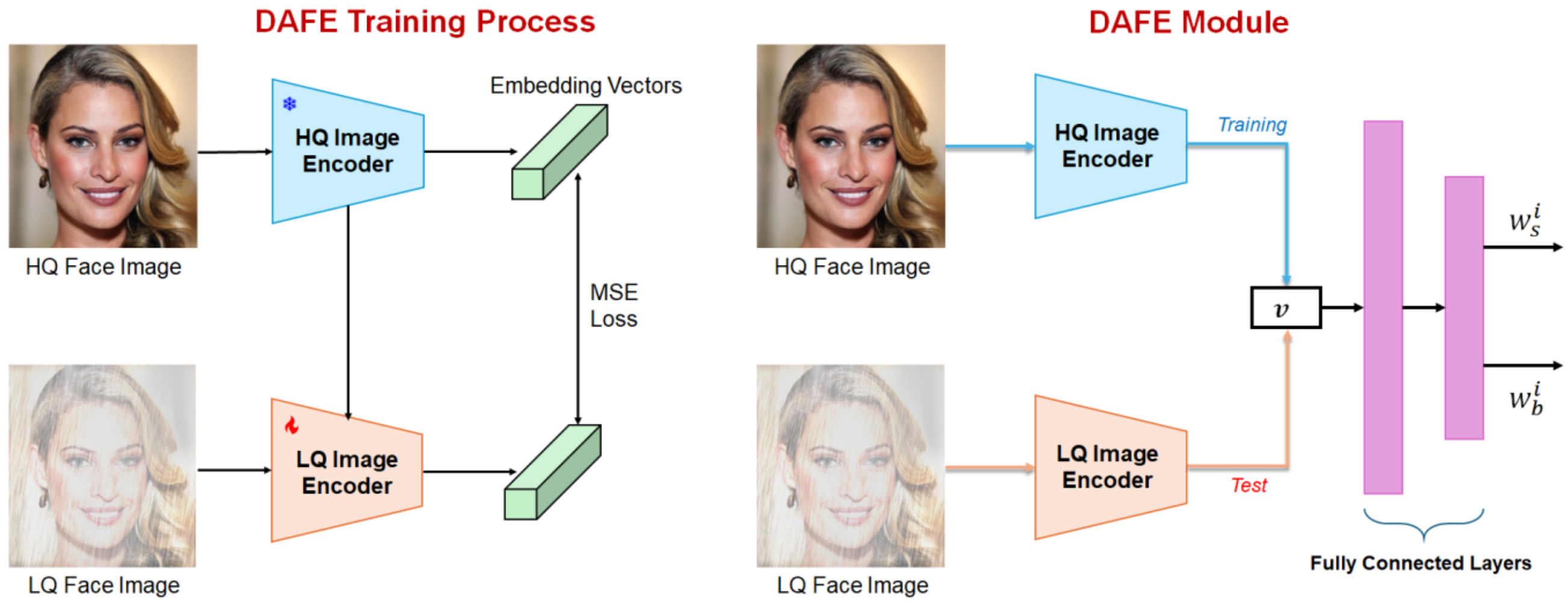

Fig. 6. DAFE training process and module.

### 3.3. DAFE Module

***While the SFFT module is capable of learning local SFFs and applying SFT for face restoration, its performance may be constrained due to the severely degraded quality of LQ face images. To mitigate this limitation, we introduce an additional DAFE module designed to generate degradation-agnostic HQ facial features.*** As illustrated in Fig. 6, the DAFE training process is depicted on the left, while the DAFE module is shown on the right. In the DAFE training process, a pretrained HQ image encoder is utilized, and the LQ image encoder is optimized to align its embedding vectors with those of the HQ encoder. Specifically, the HQ image encoder remains fixed, while only the parameters of the LQ image encoder are updated by minimizing the Mean Squared Error (MSE) between the HQ and LQ embedding vectors. In this study, the MSE loss was used for feature alignment, due to its simplicity and efficiency. In other words, contrastive loss was not employed; however, it could certainly be used as an alternative to the MSE loss. The HQ and LQ encoders were based on the ViT-B/32 architecture provided by CLIP [15], and the batch size was set to 4 for the MSE calculation. Consequently, the pretrained LQ image encoder becomes capable of extracting HQ-like facial features from LQ face images, i.e., degradation-agnostic features. Although a similar approach [12] was previously introduced for image captioning under heavy rain conditions, a fundamental distinction exists; in the image captioning model,

two encoders were employed solely for feature matching, whereas in this study, the ultimate objective is to facilitate the SFT by leveraging the pretrained LQ encoder.

The DAFE module aims to extract degradation-agnostic HQ facial features from LQ face images during inference. To achieve this, DAFE operates differently during the training and test phases.

$$Training : \boldsymbol{v}_{HQ} = E_{HQ}(\boldsymbol{H}) \tag{5}$$

$$Test : \boldsymbol{v}_{HQ} = E_{LQ}(\boldsymbol{I}) \tag{6}$$

where $E_{HQ}$ and $E_{LQ}$ denote HQ and LQ face image encoders, respectively.

During training, the HQ embedding vector $\boldsymbol{v}_{HQ}$ is first extracted using the pretrained HQ face encoder, and subsequently passed through the fully connected ($FC$) layers to generate the SFF, denoted as $\boldsymbol{w}_s^i$ and $\boldsymbol{w}_b^i$, which are distinct from the previously defined SFFs, $\boldsymbol{y}_s^i$ and $\boldsymbol{y}_b^i$.

$$[\boldsymbol{w}_s^i, \boldsymbol{w}_b^i] = FC\left(FC(\boldsymbol{v}_{HQ})\right) \tag{7}$$

Here, $FC$ represents the FC layers. The HQ face encoder remains fixed, and only the FC layers are trained.

During inference, HQ facial features can be obtained using the pretrained LQ face encoder. Specifically, the LQ face image $\boldsymbol{I}$ is processed through the LQ encoder $E_{LQ}$ and FC layers, thereby generating the SFF. ***The DAFE module thus enables the extraction of HQ facial features regardless of adverse weather conditions, making the proposed SFT-based restoration process both degradation-agnostic and more accurate.***

### 3.4. Local SFFT Enhancement Module

Given two sets of SFFs, $\{\boldsymbol{w}_s^i, \boldsymbol{w}_b^i\}$ and $\{\boldsymbol{y}_s^i, \boldsymbol{y}_b^i\}$, the SFFT Enhancement module is designed to fuse them for application in SFT. The SFFT Enhancement process is formulated as follows:

$$\boldsymbol{F}^{i} = \left(\boldsymbol{y}_{s}^{i} + \boldsymbol{w}_{s}^{i}\right) \frac{\phi\left(F^{i-1}\right) - \mu\left(\phi\left(F^{i-1}\right)\right)}{\sigma\left(\phi\left(F^{i-1}\right)\right)} + \left(\boldsymbol{y}_{b}^{i} + \boldsymbol{w}_{b}^{i}\right) \tag{8}$$

Here, it is evident that the fusion is achieved by adding two types of SFFs. Specifically, the reference standard deviation $\boldsymbol{y}_{s}^{i}$ is added to $\boldsymbol{w}_{s}^{i}$, while the reference mean $\boldsymbol{y}_{b}^{i}$ is added to $\boldsymbol{w}_{b}^{i}$. This fusion approach enhances SFT performance by facilitating degradation-agnostic feature extraction. As a result, the quality of the proposed DA-SFFT FIR model is improved.

### 3.5. Model Objectives

Following previous FIR studies [5], we adopt three types of loss functions: reconstruction loss, adversarial loss, and semantic-aware style loss. The reconstruction loss is defined as follows:

$$\mathcal{L}_{rec} = \left\|\boldsymbol{H} - \widehat{\boldsymbol{H}}\right\|_{2}^{2} + \sum_{s} \sum_{k=1}^{4} \left\|D_{s}^{k}(\boldsymbol{H}^{s}) - D_{s}^{k}(\widehat{\boldsymbol{H}}^{s})\right\|_{2}^{2} \tag{9}$$

where the first and second terms represent the mean square errors in the pixel and feature spaces, respectively. The first term ensures that the predicted HQ face image $\widehat{\boldsymbol{H}}$ closely approximates the ground truth $\boldsymbol{H}$. In the second term, $D(\cdot)$ denotes multiscale discriminators, where $s \in \{1, \frac{1}{2}, \frac{1}{4}\}$ denotes the downscale factors and $k$ indexes the layer within the discriminator. This term enforces feature-level consistency by aligning the discriminator features of $\widehat{\boldsymbol{H}}$ and $\boldsymbol{H}$.

The adversarial loss is known to be effective in preventing overly smooth outputs while enhancing fine details and generating realistic textures. In this study, the hinge loss is adopted for adversarial learning, as follows:

$$\mathcal{L}_G = \sum_s -\mathbb{E}\left(D_s(\widehat{\boldsymbol{H}}^s)\right) \tag{10}$$

$$\mathcal{L}_D = \sum_s \left\{\mathbb{E}\left[\max\left(0, 1 - D_s(\boldsymbol{H}^s)\right)\right] + \mathbb{E}\left[\max\left(0, 1 + D_s(\widehat{\boldsymbol{H}}^s)\right)\right]\right\} \tag{11}$$

where $\mathcal{L}_G$ and $\mathcal{L}_D$ denote the loss functions for the generator and the multiscale discriminators, respectively. $\mathcal{L}_G$ is designed to deceive the discriminators by making the restored HQ face image appear as real. Meanwhile, $\mathcal{L}_D$ ensures that the multiscale discriminators effectively distinguish between the predicted HR face image and the ground truth.

The semantic-aware style loss utilizes the Gram matrix to enhance texture detail recovery and is formulated, as follows:

$$\mathcal{L}_s = \sum_{i=3}^{5} \sum_{j=0}^{18} \left\| \mathcal{G}\left(\varphi_i(\widehat{\boldsymbol{H}}), M_j\right) - \mathcal{G}\left(\varphi_i(\boldsymbol{H}), M_j\right) \right\|_2^2 \tag{12}$$

where $\varphi_i$ and $M_j$ represent the VGG19 features of $ith$ layer and the parsing matrix corresponding to the $jth$ facial component, respectively. $\mathcal{G}$ denotes the gram matrix, which helps maintain natural-looking facial structures by computing feature correlations. Further details regarding the role of $\mathcal{L}_s$ can be found in [44].

The final loss function for the generator is defined, as follows:

$$\mathcal{L}_{GEN} = \lambda_s \mathcal{L}_s + \lambda_{rec} \mathcal{L}_{rec} + \lambda_G \mathcal{L}_G \tag{13}$$

where $\lambda$ represents the weighting coefficients for each loss function. The proposed DA-SFFT model is trained by alternatively minimizing $\mathcal{L}_{GEN}$ and $\mathcal{L}_D$ to achieve an optimal solution. The parameters $\lambda$ were set to be the same as the ones in [5].

## 4. Experimental Results

### 4.1. Blind Weather-Related Degradation Model

To model adverse weather conditions, this study employs a heavy rain degradation (HRD) model as it can regulate parameters related to haze effects (also referred to as rain accumulation), rain streaks, noise, blurring, and downsampling, thereby making it the most comprehensive and complex of all degradation models. It inherently encompasses both rain and haze conditions. Moreover, conducting experiments under all adverse weather scenarios is impractical; therefore, this study focuses exclusively on the HRD model for experimentation. Notably, the HRD model follows a blind approach in which the degradation parameters are randomly generated and remain unknown during the restoration process. The *blind* HRD model is defined as follows:

$$\mathbf{I} = \mathbf{T} \odot \left( \left(\mathbf{H} \otimes \mathbf{K}_{\varrho}\right) \downarrow_{s} + \sum_{i=1}^{m} \mathbf{S_i} \right) + (\mathbf{1} - \mathbf{T}) \odot \mathbf{A} \tag{14}$$

where $\mathbf{K}_{\varrho}$ denotes the blur filter with parameter $\varrho$, $\otimes$ represents the convolution operator, and $\downarrow_{s}$ is the downsampling operator with a scale factor of $s$. $\mathbf{T}$ is the transmission introduced by the scattering process of tiny water particles, $\mathbf{A}$ represents the atmospheric light, and $\mathbf{S_i}$ denotes the rain layer containing rain streaks. Additionally, $\mathbf{1}$ is a matrix of ones, and $\odot$ denotes element-wise multiplication. The rain layer $\mathbf{S_i}$ is generated by first creating a noise map and then applying a motion filter. The transmission map $\mathbf{T}$ is computed using the equation $\mathbf{T} = exp(-\beta/\boldsymbol{D})$ where $\boldsymbol{D}$ is the predicted depth map [45]. All hyperparameters, including $\varrho$, $s$, and $\beta$, are randomly selected. More details are provided in the appendix.

### 4.2. Dataset

In this study, the widely used CelebA-HQ dataset [46] is adopted for blind FIR experiments. A total of 18,000 and 5,000 HQ face images were randomly selected from CelebA-HQ to construct the training and test datasets, respectively. The selected images are resized to 512×512 using bilinear interpolation and serve as the ground-truth HQ images. The corresponding low-quality (LQ) face images are synthesized using the blind HRD model described in Eq. (14). Additionally, each image in the CelebA-HQ dataset is accompanied by a parsing map of facial components.

### 4.3. Training Details

To update the model's parameters, the Adam optimizer [47] was used. The learning rates for the generator and discriminator were set to 0.0001 and 0.0004, respectively. The training batch size was set to 4. The proposed model was implemented using PyTorch and trained on an A6000 GPU.

### 4.4. Image Quality Metrics

For quantitative image quality evaluation (IQE), both reference-based and no-reference-based metrics were considered in this study. Peak Signal-to-Noise Ratio (PSNR) and Structural Similarity (SSIM) [48] were used as reference-based IQE metrics to assess similarity in terms of pixel accuracy and structural reconstruction. However, these metrics tend to favor smooth textures and may not fully align with human visual perception. To evaluate perceptual image quality, the Learned Perceptual Image Patch Similarity (LPIPS) [49] and Fréchet Inception Distance (FID) [50] were also employed, as they do not require ground-truth images. These metrics measure the statistical distance between the restored HQ face images and the ground-truth HQ images. For PSNR and SSIM, higher values indicate better restoration quality, whereas for FID and LPIPS, lower values indicate higher restoration quality.

### 4.5. Visual Quality Comparison

In this section, we evaluate the proposed DA-SFFT model alongside conventional state-of-the-art (SOTA) methods to assess their restoration quality. Following the evaluation protocol of PSFRGAN [5], we select representative SOTA blind FIR models, including HiFaceGAN [51], PSFRGAN [5], and VQFR [52]. We also include general image restoration models based on GAN and diffusion frameworks. Specifically, we consider Heavy Rain Removal (HRR) [32], Real-ESRGAN [53], and BFRffusion [54] in our evaluation. While we experimented with other models such as Uformer[55], ESRGAN[56], SRGAN [40], and DifFace[30], they are excluded from this main text due to their subpar restoration quality. Detailed results are provided in the Appendix. Furthermore, we compare our FIR model with TANet [57], a recent all-in-one restoration model designed for adverse weather conditions, demonstrating that our model clearly outperforms TANet.

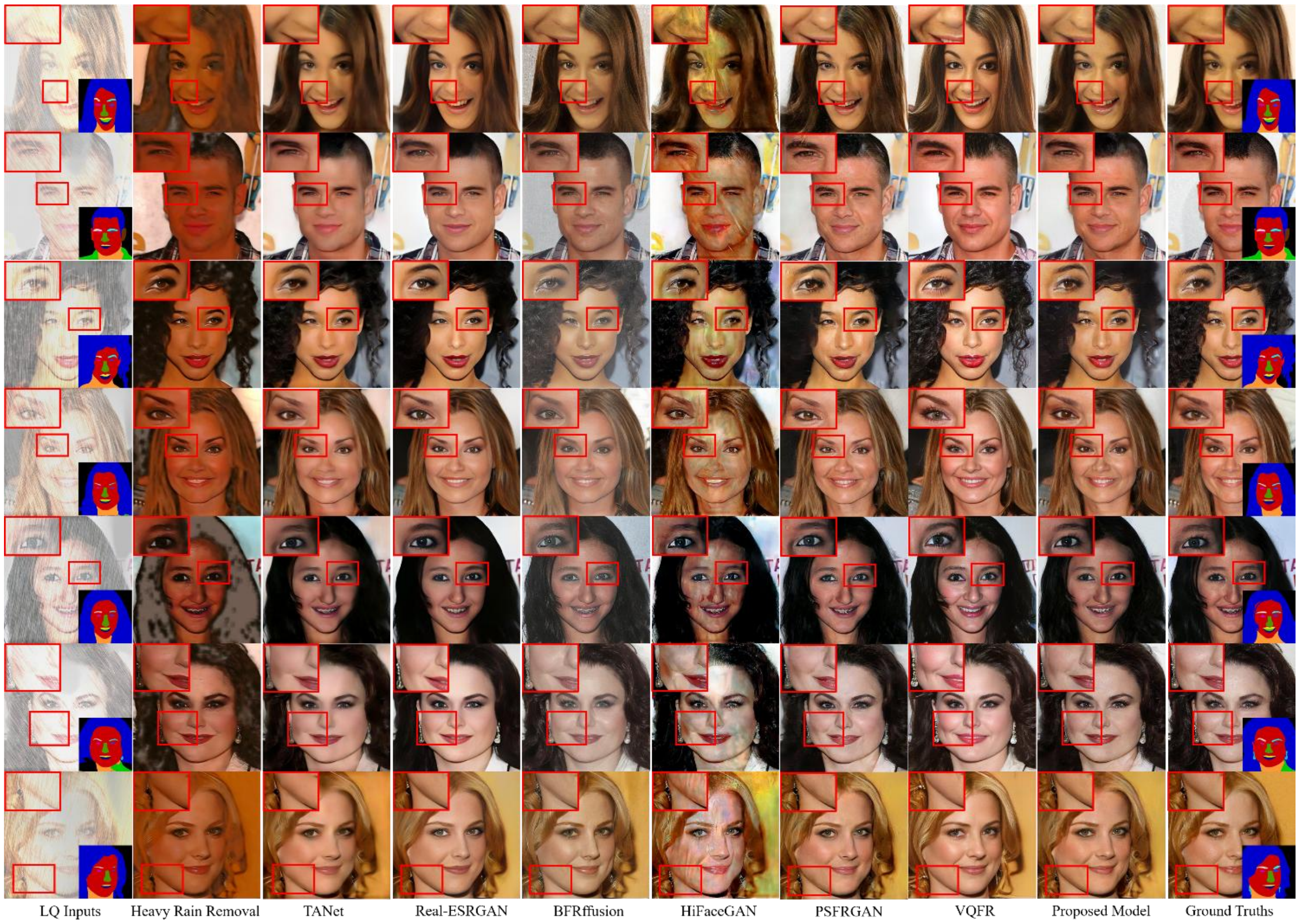

Fig. 7. Experimental results. From left to right: LQ images, HRR [32], TANet [57], Real-ESRGAN [53], BFRffusion [54], HiFaceGAN [51], PSFRGAN [5], VQFR [52], the proposed DA-SFFT model, and ground-truths images. (We recommend downloading and zooming in on the high-resolution image to clearly observe the differences in restoration quality.)

Fig. 7 shows the restored images generated by the aforementioned models. As observed in the figure, blind FIR models generally outperform general image restoration models due to their architectures specifically optimized for facial restoration. Although Real-ESRGAN and BFRffusion are considered SOTA among general image restoration methods, they still show notable limitations: Real-ESRGAN tends to over-smooth fine details, while BFRffusion produces diluted colors compared to the ground-truth images. In addition, while TANet is designed for unified image restoration under adverse weather conditions, its architecture is not optimized for highly structured facial images, resulting in overly smoothed textures.

Blind FIR models, except for HiFaceGAN, are successful in producing realistic and faithful facial restorations. However, in the last two rows, the VQFR model distorts the shape of the earrings, making them noticeably different from the originals. This indicates that VQFR tends to prioritize realistic appearance over strict fidelity to the input. Meanwhile, both PSFRGAN and the proposed model achieve faithful face restoration. However, PSFRGAN is less effective than the proposed model in removing artifacts and reconstructing facial shapes and details. Specifically, in the second and third rows, the proposed DA-SFFT method provides a more accurate depiction of eye shapes. Additionally, the proposed model demonstrates superior capability in eliminating texture distortions, as evidenced by the first three rows. The proposed DA-SFFT model is an advanced version of PSFRGAN, enhanced by the integration of the DAFE, SFFT, and SFFT enhancement modules for improved facial representation. Compared to PSFRGAN, the proposed model demonstrates clear improvements in reconstructing facial shapes and fine details around the eyes, lips, and nose, owing to the contributions of the DAFE and local SFFT modules.

Table 1. Quantitative evaluations.

| **Model types** | **Metrics** / **Models** | **PSNR↑** | **SSIM↑** | **LPIPS↓** | **FID↓** |
|---|---|---|---|---|---|
| HRR model | HRR [32] | 14.5337 | 0.6087 | 0.3106 | 39.8703 |
| SR model | Real-ESRGAN [53] | 22.5930 | 0.7173 | 0.2093 | 25.9977 |
| All-in-one model | TANet [57] | 23.1651 | 0.7121 | 0.2111 | 37.4778 |
| Diffusion model | BFRffusion [54] | 18.5903 | 0.5800 | 0.2399 | 12.5220 |
| FIR models | HiFaceGAN [51] | 20.3458 | 0.5925 | 0.2688 | 71.3781 |
| | PSFRGAN [5] | 22.2651 | 0.6814 | 0.2317 | 18.6487 |
| | VQFR [52] | 21.3815 | 0.6143 | 0.2086 | 13.1778 |
| | **Proposed DA-SFFT Model** | **23.5901** | **0.7205** | **0.2044** | **11.8186** |

### 4.6. Quantitative Evaluation

Quantitative results are presented in Table 1, comparing the proposed DA-SFFT model with various restoration approaches. Among the general restoration methods, TANet and Real-ESRGAN achieve the highest PSNR and SSIM, respectively, while BFRffusion records the lowest FID, indicating relatively better perceptual quality. However, these models still fall short compared to specialized FIR models in restoring fine facial details. Within the FIR category, the proposed model outperforms all others across all evaluation metrics. It achieves the highest PSNR and SSIM, and the lowest LPIPS and FID, demonstrating superior fidelity and perceptual quality. In contrast, HiFaceGAN exhibits the weakest performance among FIR models, particularly with the highest FID, suggesting poor alignment with real facial distributions. Compared to PSFRGAN and VQFR, the proposed DA-SFFT model shows consistent improvements, validating the effectiveness of the DAFE and local SFFT modules in enhancing both structural accuracy and visual realism.

Table 2. Effectiveness of the SFFT and DAFE modules.

| **Models** | **PSNR↑** | **SSIM↑** | **LPIPS↓** | **FID↓** |
|---|---|---|---|---|
| **PSFRGAN [5]** | 22.2651 | 0.6814 | 0.2317 | 18.6487 |
| **Proposed Model (SFFT)** | **22.5300** | **0.6989** | **0.2160** | **13.7977** |
| **Proposed DA-SFFT Model (SFFT + DAFE)** | **23.5901** | **0.7205** | **0.2044** | **11.8186** |

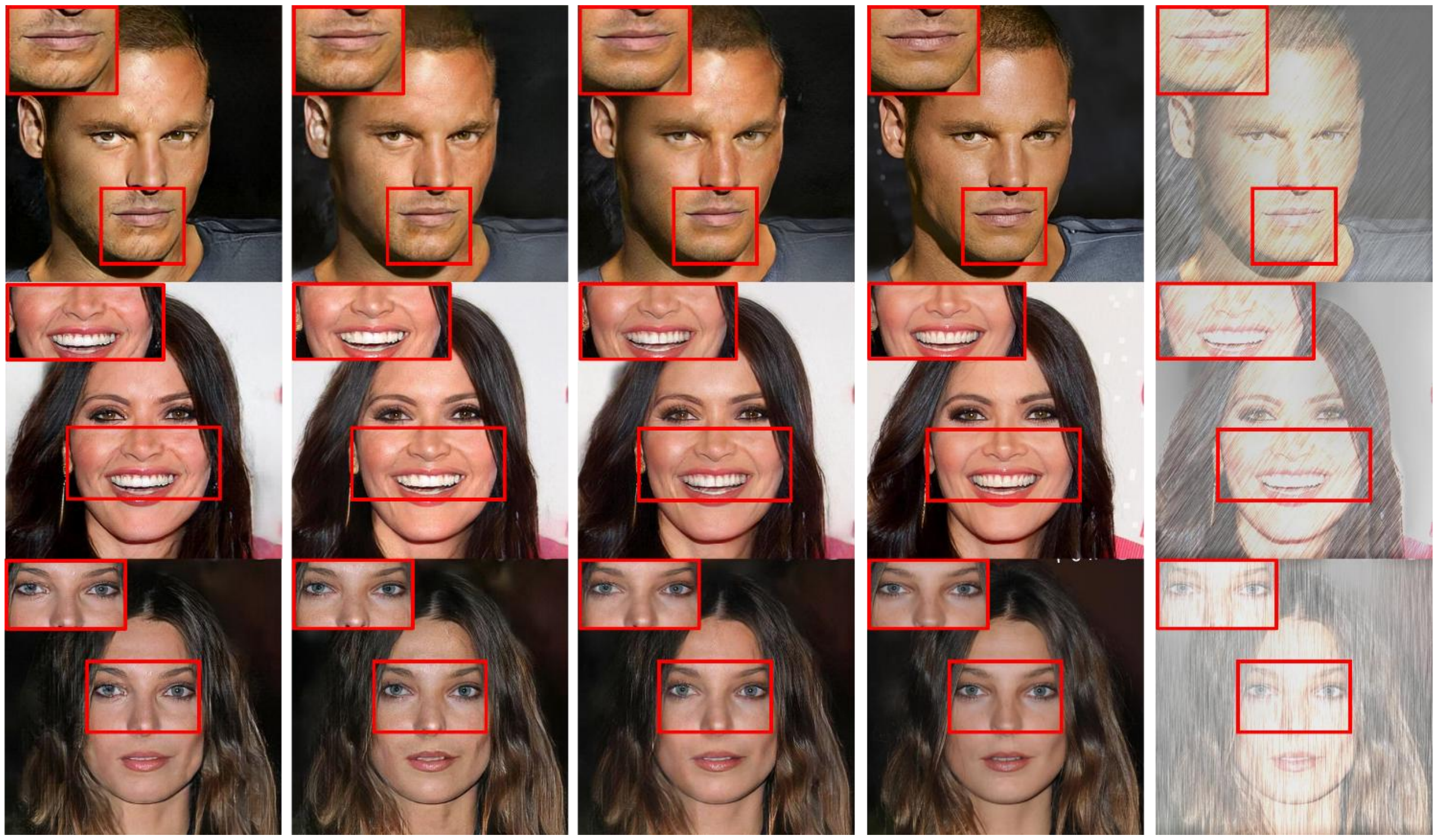

Fig. 8. Visual quality comparison: PSFRGAN, proposed model (SFFT), proposed DA-SFFT model (SFFT+DAFE), ground truths, and LQ inputs (from left to right).

### 4.7. Ablation Study

#### 4.7.1. Evaluation of the SFFT and DAFE modules

Table 2 presents an ablation study that evaluates the effectiveness of the SFFT and DAFE modules. Compared to PSFRGAN, the proposed model with only the SFFT module shows noticeable improvements in all metrics, including higher PSNR and SSIM, and lower LPIPS and FID demonstrating the benefits of the local SFFT mechanism. Specifically, it enables the local statistical distributions of LQ facial components to be converted into those of their HQ counterparts, thereby enhancing the representation of local facial features. When both the SFFT and DAFE modules are incorporated, the proposed DA-SFFT model achieves the best performance across all metrics, recording the highest PSNR, SSIM, and the lowest LPIPS and FID. These results confirm that the DAFE module further enhances structural fidelity by extracting HQ face features regardless of adverse weather conditions, thereby making the proposed restoration process more degradation-agnostic.

Overall, the two modules work synergistically, leading to significant improvements in both structural accuracy and perceptual quality in face restoration.

Fig. 8 demonstrates the effectiveness of the local SFFT and DAFE modules in removing image artifacts and restoring fine details and facial structures. In the first and second columns, the local SFFT module effectively reduces texture distortions around the jaw and eye areas, compared to PSFRGAN. It also corrects lip color to better match the ground truth and alleviates structural distortions in the teeth. In the third column, the DAFE module more accurately reconstructs the beard and refines the shape of the teeth. Additionally, the skin is rendered with smoother, more natural textures. These results indicate that both modules contribute to enhancing structural accuracy and perceptual quality in face restoration.

#### 4.7.2. Feature visualization

To verify whether the LQ features learned by the LQ encoder can be aligned with the HQ features, we applied t-SNE [58], a dimensionality reduction technique used to visualize high-dimensional data in a low-dimensional space. Fig. 9 presents the HQ feature and its corresponding LQ features before and after applying the aligned LQ encoder. In the figure, the HQ feature marked with a blue circle represents the embedding result of the HQ face image passed through the HQ encoder. The LQ features, indicated by a triangle and a star, represent the embedding results before and after alignment through the LQ encoder, respectively. As shown in the figure, the LQ feature after alignment via the LQ encoder is positioned closer to the HQ feature compared to the LQ feature before alignment. This visualization confirms that the DAFE module can extract HQ-like features from LQ inputs, making the LQ features more degradation-agnostic.

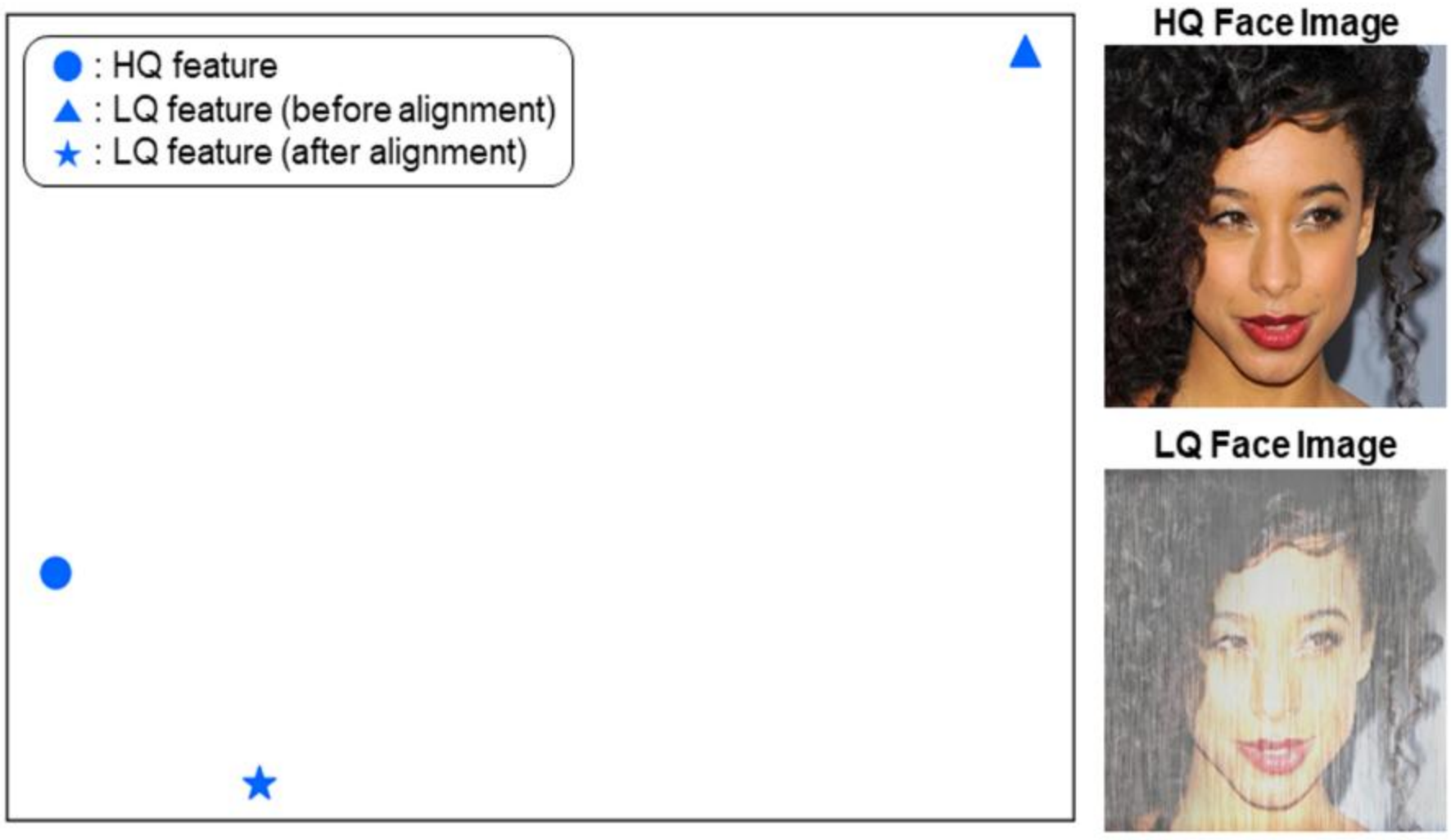

Fig. 9. t-SNE visualization of features: HQ feature and its corresponding LQ feature before and after applying the aligned LQ encoder.

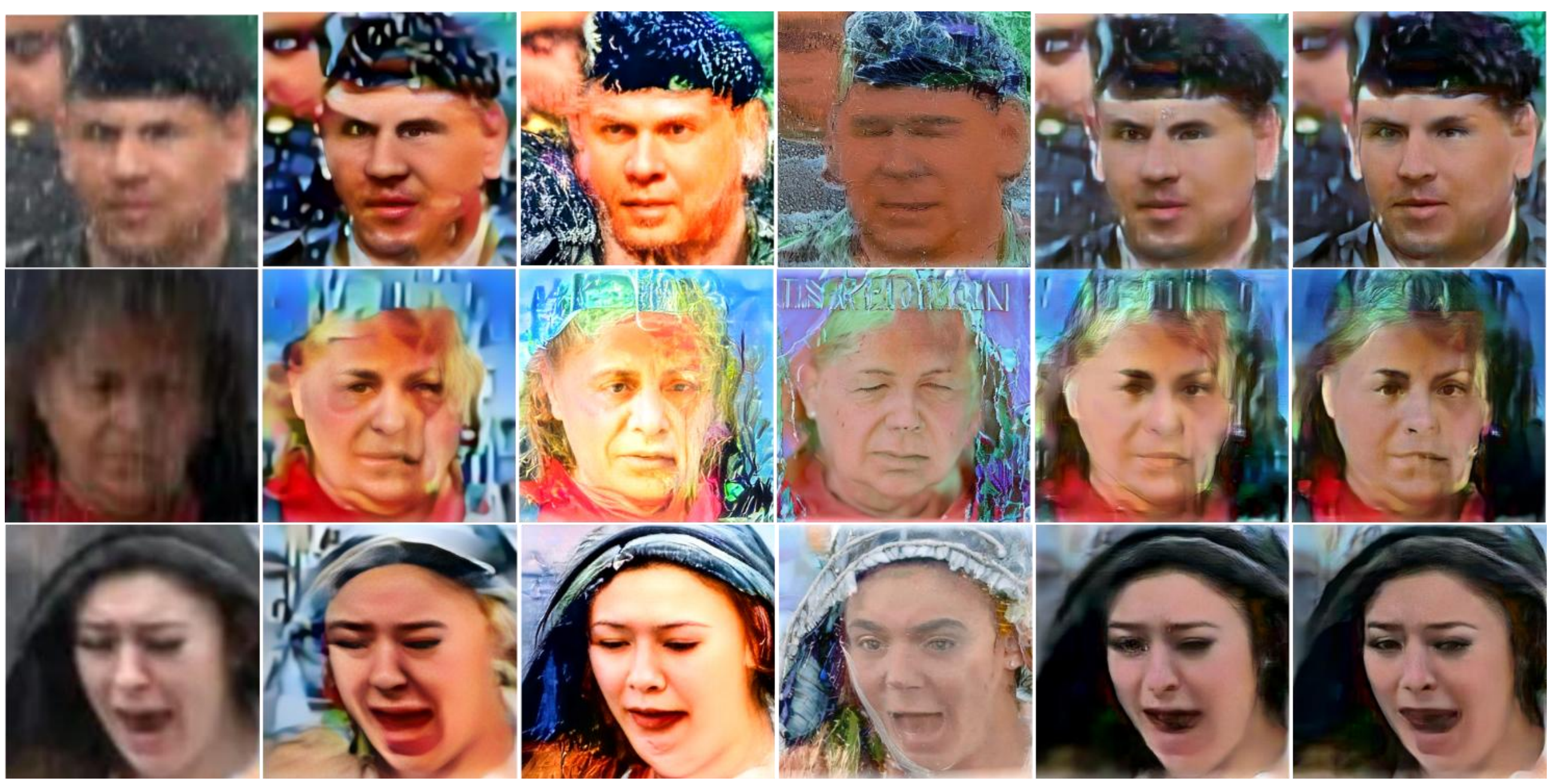

Fig. 10. Application to real-world face images. From left to right: LQ inputs, Real-ESRGAN, VQFR, BFRffusion, PSFRGAN, and the proposed model. (We recommend downloading and zooming in on the high-resolution image to clearly observe the differences in restoration quality.)

### 4.8. Discussion and limitations

It is worthwhile to examine whether both the conventional and proposed FIR models perform effectively on LQ images captured under real-world weather conditions. Fig. 10 presents examples of restored face images generated by Real-ESRGAN, VQFR, BFRffusion, PSFRGAN, and the proposed method. Under real-world weather conditions, the LQ images suffered from multiple degradations, including severe resolution loss and noticeable rain streaks.

As shown in Fig. 10, the BFRffusion model produced unnatural texture patterns, particularly in the background, and the reconstructed facial appearances deviated noticeably from those in the original LQ images. The Real-ESRGAN and VQFR models distorted the shapes of facial components and background regions, respectively, leading to overall poor visual quality. While PSFRGAN better preserved the shape of facial components, it still lacked sharpness in critical regions such as the lips, beard, and nose. In contrast, the proposed DA-SFFT model achieved sharper representations of facial features, particularly in the beard, lips, and eye regions. Specifically, it produced a more accurate and clearer depiction of the lips in the first row, and noticeably sharper and more precise eye regions in the second and third rows. This highlights the effectiveness of the DAFE and SFFT modules in enhancing facial structure even under adverse weather conditions.

Table 3. No-reference quantitative evaluation for real-world images.

| Metrics | Real-ESRGAN | VQFR | BFRffusion | PSFRGAN | Proposed Model |
|---|---|---|---|---|---|
| LPIPS | 0.3419 | 0.4232 | 0.4440 | 0.2914 | **0.2875** |
| | 0.6356 | 0.8674 | 0.8002 | 0.6681 | **0.6063** |
| | 0.5241 | 0.6511 | 0.6468 | 0.4048 | **0.3572** |
| FID | 327.3969 | 315.0418 | 239.1983 | 283.5627 | **216.9697** |
| | 278.9465 | 405.7689 | 425.9391 | 381.5545 | **230.1088** |
| | 339.7229 | 300.7581 | 429.1472 | 308.4457 | **266.2595** |

Table 3 presents the no-reference quantitative evaluation of the results produced by each model for the three LQ input images shown in Fig. 10. The FID and LPIPS scores in each row of Table 3 correspond to the results of the conventional SOTA models for the real-world face images in the corresponding rows of Fig. 10. As expected, the proposed model achieved the best performance in terms of LPIPS and FID. Although the overall quality of the reconstructed face image remains constrained due to domain shift issues, ***the results demonstrate that the proposed FIR model is capable of restoring critical facial components from visually unrecognizable facial images, thereby highlighting its potential to enable reliable face recognition under adverse weather conditions.***

To complement the limitations of the proposed model, two potential directions can be considered. First, prompt learning can serve as an effective strategy to achieve a weather-adaptive restoration process. In future work, we plan to incorporate additional types of degraded face images, such as those affected by snow or low-light conditions. Inspired by recent all-in-one restoration approaches [59, 41], weather-related text vector prompts can be learned through contrastive learning with corresponding degraded facial images. The learned text prompts will then be integrated into the proposed generator to enable adaptive face restoration under various weather conditions. Second, local DAFE modules will be explored in future research. In the current version, a global DAFE module was designed, where degradation-agnostic features are extracted from entire face images via the pretrained LQ image encoder. However, this global DAFE has limited capability to represent fine-grained facial components. Therefore, local DAFE modules will be employed to capture degradation-agnostic representations for specific facial regions (e.g., eyes, nose, and mouth), thereby improving the fidelity and sharpness of the restored face images.

In addition, our model has a parameter count of 289.1 M, a model size of 1102.87 MB, and an inference speed of 0.078 sec. The model is slightly larger due to the inclusion of the DAFE encoder and facial attention module. However, under adverse weather conditions, achieving high-quality face restoration sufficient for face recognition is of greater importance, as demonstrated in the real-world experiments shown in Fig. 10. Therefore, the proposed FIR model is valuable in its ability to restore critical facial components from visually unrecognizable images, highlighting its potential to enable reliable face recognition under challenging weather conditions.

## 5. Conclusions

This paper presents a novel blind FIR model designed to handle adverse weather conditions. The proposed DA-SFFT model incorporates two key modules to effectively remove weather-induced artifacts and reconstruct facial structures with higher fidelity. First, the local SFFT module transforms the local statistical distributions of facial components in LQ images to resemble those in HQ images, thereby enhancing both color consistency and structural representation. Second, to address the limitations of the local SFFT module—particularly under conditions of low visibility and heavy rain, which impede reliable feature extraction—the DAFE module is introduced. This module enables the extraction of degradation-agnostic SFF features by aligning feature representations from HQ and LQ image encoders, allowing the SFFT-based restoration process to adapt robustly to weather-related degradations. ***Experimental results demonstrate that the proposed FIR model is capable of restoring critical facial components from visually unrecognizable facial images, thereby highlighting its potential to enable reliable face recognition under adverse weather conditions. Furthermore, the proposed DA-SFFT model outperforms state-of-the-art GAN- and diffusion-based FIR and IR methods, particularly in suppressing texture distortions and accurately reconstructing facial structures.*** The effectiveness of both the local SFFT and DAFE modules is also empirically validated.

**Funding:** This research was supported by 2025 Fostering project on Regional Characterization Program through the INNOPOLIS funded by Ministry of Science and ICT (RS-2025-02312252)

# Appendix

### A. Details of weather-related blind degradation model

As described in the paper, the degradation model is expressed as follows:

$$\mathbf{I} = \mathbf{T} \odot \left( \left( \mathbf{H} \otimes \mathbf{K}_{\varrho} \right) \downarrow_{s} + \sum_{i=1}^{m} \mathbf{S_i} \right) + (\mathbf{1} - \mathbf{T}) \odot \mathbf{A} \quad \text{(A. 1)}$$

where

- $\mathbf{K}_{\varrho}$ is the blur filter. A Gaussian blur filter is applied with the standard deviation (which controls the strength of blurring) randomly selected from the uniform distribution $\varrho \sim U(0, 2.5)$.
- The operator $\downarrow_s$ represents downsampling, with the scale factor s randomly selected from the uniform distribution $s \sim U(32, 256)$.
- To generate the rain layer $\mathbf{S_i}$, a noise map is first created according to a Gaussian distribution, with the mean $\mu \sim U(-1, -0.8)$ and the standard deviation $\sigma \sim U(0.7, 1.0)$. A motion filter is then applied to the noise map. The motion parameters, including the length $\ell = 45$ and angle $\theta = \{55, 80, 90, 110, 125\}$, are set accordingly.
- The transmission map **T** is computed using the equation $\mathbf{T} = exp(-\beta \cdot \mathbf{D}_n)$ where the parameter β is randomly selected from the uniform distribution $\beta \sim U(2.6, 4.6)$, and $\mathbf{D}_n = (1/(\mathbf{D} + 0.1))$ is the reciprocal of the depth map **D**, normalized to the range [0,1].
- Finally, the atmospheric light **A** is randomly selected from the range [0.1, 0.8].

### B. Visual quality of other models

As discussed in the main body, other image restoration models-including SRGAN, ESRGAN, Uformer, and DifFace-were also tested. As shown in Fig. A. 1, the visual quality produced by these models is inferior to that of the proposed method. Specifically, they exhibit poor overall sharpness along with noticeable color and texture distortions.

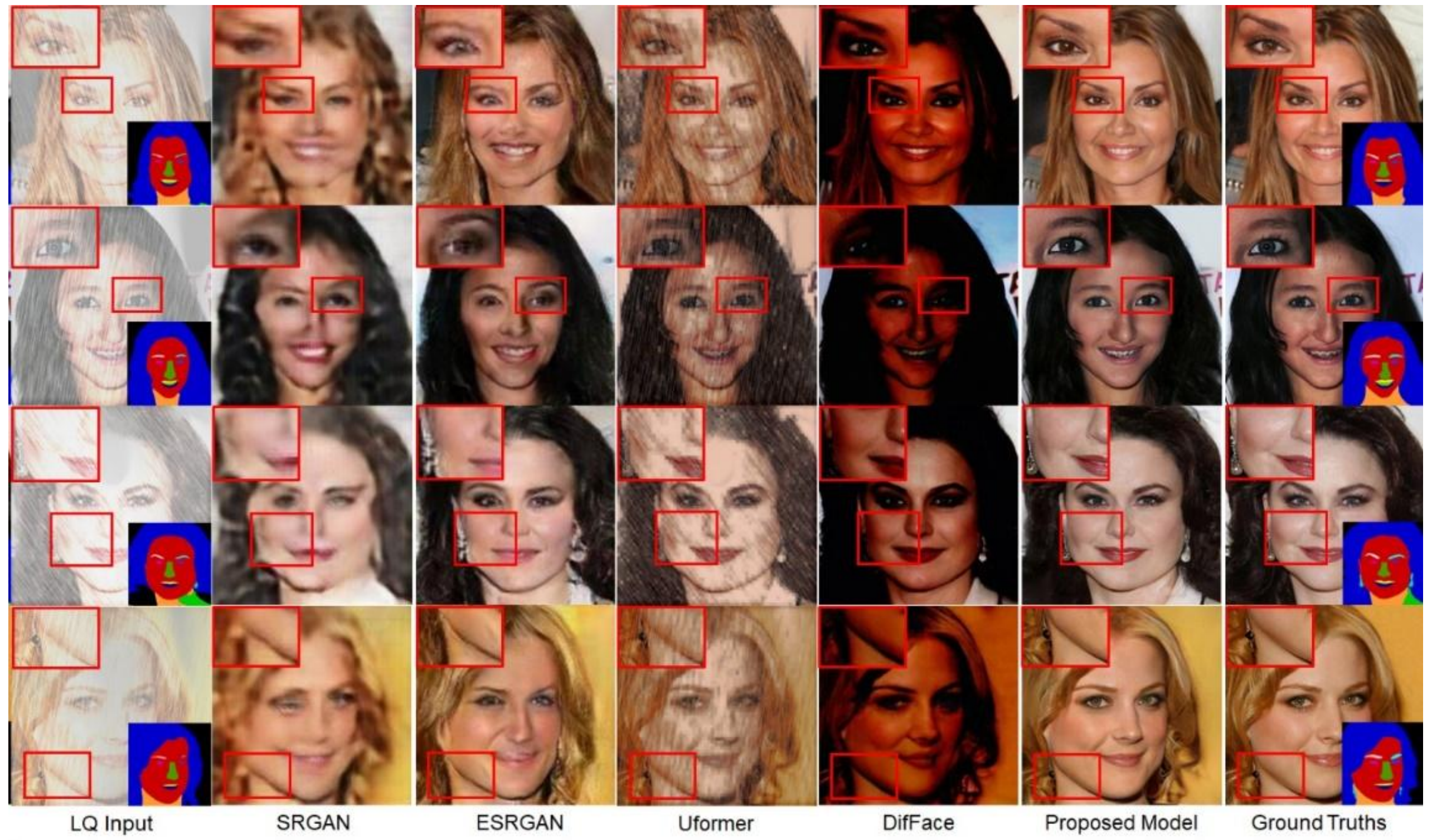

Fig. A. 1. Experimental results. LQ images, SRGAN [40], ESRGAN [56], Uformer [55], DifFace [30], the proposed model, and ground truths (from left to right).